\pdfoutput=1
\documentclass[10pt,twocolumn,letterpaper]{article}

\usepackage[pagenumbers]{cvpr} 
\usepackage[accsupp]{axessibility}
\usepackage{graphicx}
\usepackage{amsmath}
\usepackage{amssymb}
\usepackage{booktabs}
\usepackage{color}
\usepackage{xcolor}
\usepackage{multirow}
\usepackage{makecell}

\usepackage{colortbl}

\definecolor{maroon}{cmyk}{0,0.87,0.68,0.32}

\usepackage[pagebackref,breaklinks,colorlinks]{hyperref}

\usepackage[capitalize]{cleveref}
\crefname{section}{Sec.}{Secs.}
\Crefname{section}{Section}{Sections}
\Crefname{table}{Table}{Tables}
\crefname{table}{Tab.}{Tabs.}

\def\confName{CVPR}
\def\confYear{2022}

\def\etal{\emph{et~al.}} 
\def\ie{\textit{i.e.}}
\def\eg{\textit{e.g.}}

\begin{document}

\title{Multi-View Consistent Generative Adversarial Networks \\ for 3D-aware Image Synthesis}
\author{Xuanmeng Zhang$^{1,2}$ \thanks{This work was done during an internship at Alibaba.}
\quad Zhedong Zheng$^{1}$  \quad Daiheng Gao$^{2}$  \\
Bang Zhang$^{2}$  \quad Pan Pan$^{2}$ \quad Yi Yang$^{3}$ \\
$^1${ReLER, AAII, University of Technology Sydney} \\
\quad $^2${DAMO Academy, Alibaba Group} \quad $^3${Zhejiang University} \\
{\tt\small \{zhangxuanmeng.zxm, zdzheng12\}@gmail.com} \\
{\tt\small \{daiheng.gdh, zhangbang.zb, panpan.pp\}@alibaba-inc.com yangyics@zju.edu.cn}
}

\maketitle

\begin{abstract}
3D-aware image synthesis aims to generate images of objects from multiple views by learning a 3D representation.
However, one key challenge remains: 
existing approaches lack geometry constraints, hence usually fail to generate multi-view consistent images. 
To address this challenge,  we propose  \textbf{M}ulti-\textbf{V}iew \textbf{C}onsistent \textbf{G}enerative \textbf{A}dversarial \textbf{N}etworks (\textbf{MVCGAN}) for high-quality 3D-aware image synthesis with geometry constraints.
By leveraging the underlying 3D geometry information of generated images, \ie, depth and camera transformation matrix, we explicitly establish stereo correspondence between views to perform multi-view joint optimization.
In particular, we enforce the photometric consistency between pairs of views and integrate a stereo mixup mechanism into the training process, encouraging the model to reason about the correct 3D shape.
Besides, we design a two-stage training strategy with feature-level multi-view joint optimization to improve the image quality.
Extensive experiments on three datasets demonstrate that MVCGAN achieves the state-of-the-art performance for 3D-aware image synthesis. 
\end{abstract}

\begin{figure}
{
      \includegraphics[width=0.4\linewidth]{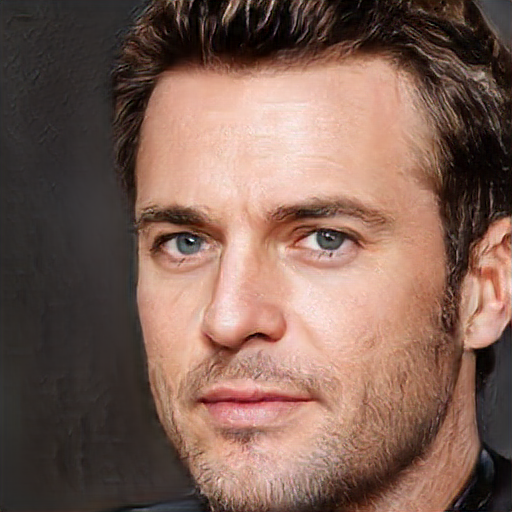}
      \vspace{-0.27mm}
      \hspace{-1.275mm}
      \includegraphics[width=0.6\linewidth]{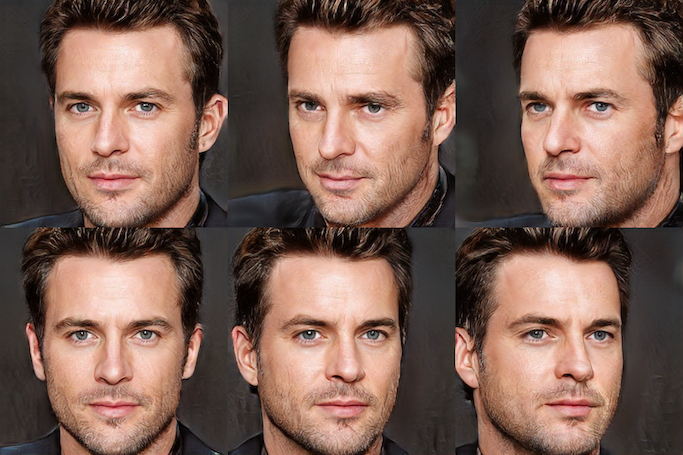}
      \includegraphics[width=0.4 \linewidth]{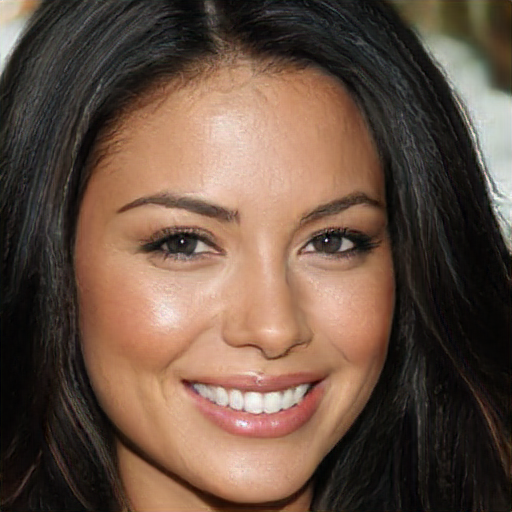}
      \hspace{-1.275mm}
      \includegraphics[width=0.6 \linewidth]{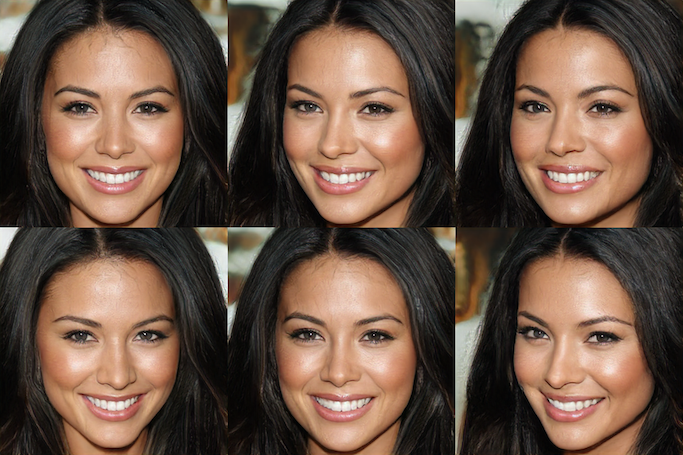}
}
\vspace{-5mm} 
  \caption{Images synthesized by MVCGAN on the CELEBA-HQ~\cite{karras2018progressive} dataset. }
 \label{fig:select}
\vspace{-5mm} 
\end{figure}

\begin{figure*}[]
 \centering
  \includegraphics[width=1.0 \linewidth]{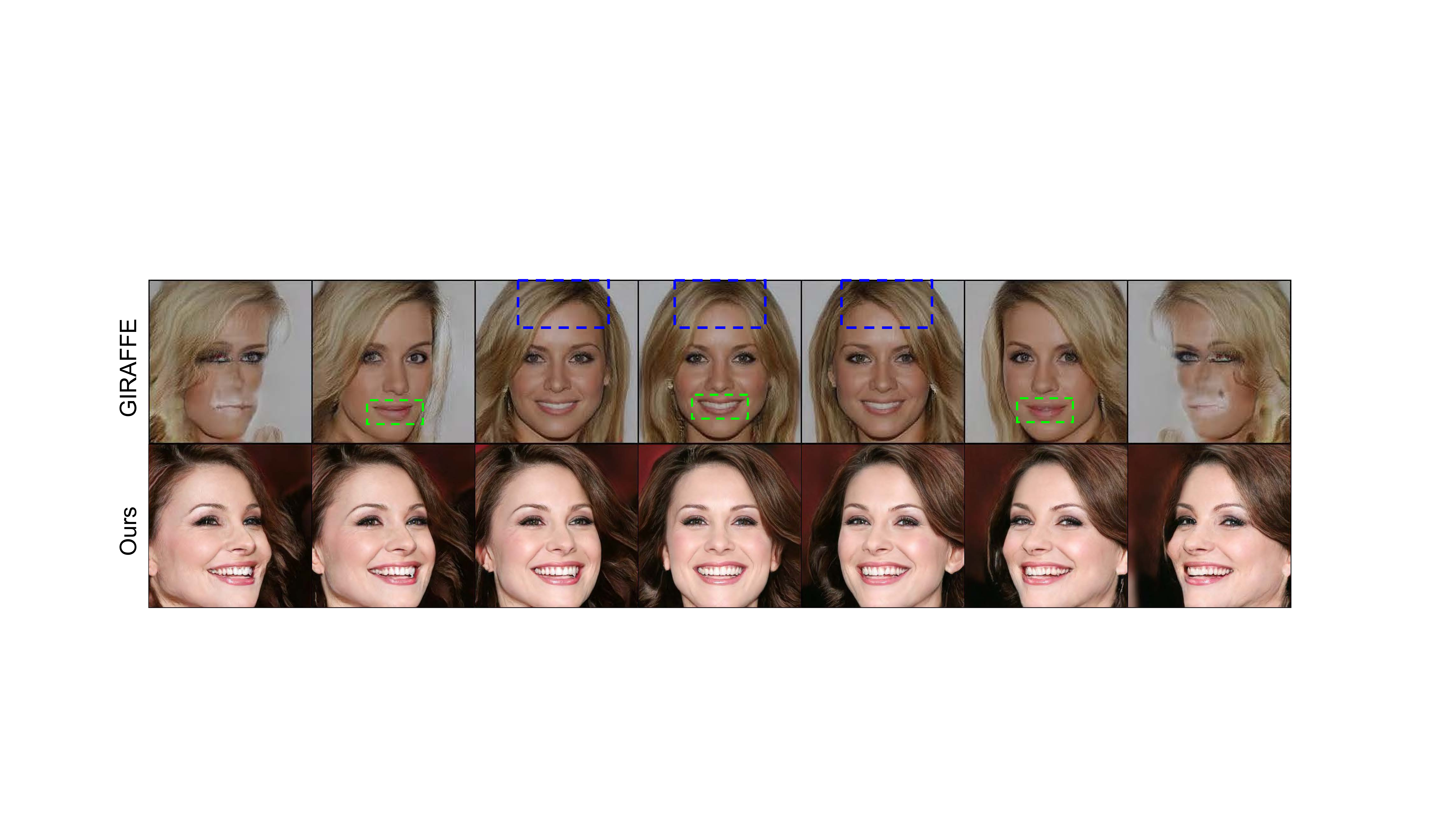}
  \vspace{-3mm}
  \caption{\textbf{Typical failure cases.}
Taking a representative method GIRAFFE~\cite{niemeyer2021giraffe} as an example,
the generated images in the first row have obvious appearance inconsistent artifacts between views, such as the direction of hair~(\textcolor{blue}{blue} box) and the opening mouth~(\textcolor{green}{green} box).
Besides, we notice that GIRAFFE~\cite{niemeyer2021giraffe} suffers from collapsed results under large pose variations~(see the leftmost and rightmost pictures in the first row),
which indicates that the model does not learn an appropriate 3D shape.
In contrast, our method generates high-quality images with multi-view consistency~(see the second row).
}
\label{fig:failture}
\vspace{-3mm}
\end{figure*}

\section{Introduction}
\label{sec:intro}
We study the problem of 3D-aware image synthesis, aiming at generating images with explicit control over the camera pose.
Generating photorealistic and editable image content is a long-standing problem in computer vision and graphics.
In the past years, generative adversarial networks (GAN)~\cite{goodfellow2014generative} have demonstrated impressive results in synthesizing
high-resolution images of high quality from unstructured image collections~\cite{brock2018large,zhu2017unpaired,choi2018stargan,karras2018progressive,huang2017arbitrary,karras2019style,zheng2019joint,karras2020analyzing,choi2020stargan}.
Despite the tremendous success, most of the methods typically only learn the manifold of 2D images while ignoring the 3D representation of the scene.

Several works consider the task of 3D-aware image synthesis~\cite{alhaija2018geometric,nguyen2019hologan,zhu2018visual,liao2020towards,nguyen2020blockgan,henderson2020leveraging,DeVries_2021_ICCV}, which can generate images of objects from multiple views by learning a 3D-aware generative model.
Different from 2D generative adversarial networks, 3D-aware image synthesis models learn 3D scene representations from images, such as voxels~\cite{nguyen2019hologan,nguyen2020blockgan}, intermediate 3D primitives~\cite{liao2020towards}, and neural radiance fields (NeRF)~\cite{schwarz2020graf,chan2021pi,niemeyer2021giraffe,DeVries_2021_ICCV}.
Among these approaches, NeRF-based approaches~\cite{schwarz2020graf,chan2021pi,niemeyer2021giraffe,DeVries_2021_ICCV} have gained a surge of interest due to the extraordinary performance for high-fidelity view synthesis.
However, one key challenge remains in existing approaches~\cite{schwarz2020graf,chan2021pi,niemeyer2021giraffe}:
they do not guarantee geometry constraints between views, 
hence usually fail to generate multi-view consistent images in some views.

In this paper, we address this problem by proposing MVCGAN, a multi-view consistent generative model for high-quality 3D-aware image synthesis with geometry constraints~(see Fig.~\ref{fig:select}). 
We first present typical failure cases of existing approach~\cite{niemeyer2021giraffe} in Fig.~\ref{fig:failture}.
Then we identify the cause of the inconsistent phenomenon between views: previous methods optimize a single view of the generated image independently while ignoring the geometry constraints between views~(see Sec.~\ref{sec:image-level multi-view}).
To tackle this problem, we take inspiration from the classical multi-view geometry methods~\cite{chen1993view,debevec1996modeling,andrew2001multiple,seitz1996view,zhou2017unsupervised,godard2019digging} and jointly optimize multiple views with geometry constraints.
By leveraging the underlying 3D geometry information,
we explicitly establish stereo correspondence between views through projective geometry.
To encourage the network to reason about the correct 3D shape, we perform multi-view joint optimization by enforcing the photometric consistency between pairs of views with re-projection loss and integrating a stereo mixup mechanism into the training process.
Therefore, the generator not only learns the manifold of 2D images, but also ensures the correctness of the underlying 3D shape.

Besides, we notice that NeRF-based generative approaches~\cite{schwarz2020graf,chan2021pi,niemeyer2021giraffe} typically struggle to render high-resolution images with fine details due to the huge computational complexity of NeRF model~\cite{mildenhall2020nerf}.
Existing methods~\cite{schwarz2020graf,chan2021pi,niemeyer2021giraffe} adopt different strategies to synthesize high-resolution images.
However, they all have limitations.
GRAF~\cite{schwarz2020graf} introduces a multi-scale patch-based discriminator, which causes uneven image quality and local overfitting to the last batch.
pi-GAN~\cite{schwarz2020graf} increases the resolution of the generator by sampling rays more densely, which still requires intensive memory consumption.
GIRAFFE~\cite{niemeyer2021giraffe} combines the 3D representation with a neural rendering pipeline, which suffers from collapsed results under large pose variations.
In this paper, we adopt a hybrid MLP-CNN architecture to disentangle the geometry of 3D shape from the fine details of 2D appearance.
In particular, the MLP-based NeRF model~\cite{mildenhall2020nerf} renders the geometry of 3D shape, and the CNN-based decoder produces fine details for 2D appearance.
The structure can generate photorealistic high-resolution images  while alleviating the computation-intensive problem.

Overall,  our contributions are summarized as follows:
\begin{enumerate}
\item We identify one challenging problem of missing the geometry constraints in 3D-aware image synthesis, which leads to inconsistent images across views. 

\item  
We propose a multi-view consistent generative model (MVCGAN) for high-quality 3D-aware image synthesis.
By establishing the geometry constraints, we jointly optimize multiple views to guarantee the geometry consistency between views.
Besides, we design a two-stage training strategy with the feature-level multi-view joint optimization to further improve the image quality. 

\item  
We demonstrate the effectiveness of the proposed approach through evaluating on various datasets, \textit{i.e.}, CELEBA-HQ~\cite{karras2018progressive}, FFHQ~\cite{karras2019style}, and AFHQv2~\cite{choi2020stargan}. 
Extensive experiments substantiate that MVCGAN  achieves the state-of-the-art performance for 3D-aware image synthesis. 

\end{enumerate}

\section{Related Work}
\label{sec:related}

\noindent \textbf{Multi-view Geometry.}
A large number of approaches reconstruct 3D structure with multi-view geometry constraints as supervision signals, such as COLMAP~\cite{schonberger2016structure} and ORB-SLAM~\cite{mur2015orb}.
In recent years, Some deep learning techniques~\cite{zhou2017unsupervised,godard2019digging,yao2018mvsnet} also combine traditional approaches~\cite{chen1993view,collins1996space,szeliski1998stereo} to address 3D vision problems.
Inspired by the classical multi-view geometry methods~\cite{chen1993view,debevec1996modeling,andrew2001multiple,seitz1996view,zhou2017unsupervised,godard2019digging},
we explicitly involve the geometry constraints in the training process for learning a reasonable 3D shape.

\noindent \textbf{Neural Radiance Fields.}
Recently, using volumetric rendering and implicit function to synthesize novel views of a scene has gained a surge of interest. 
Mildenhall~\etal~\cite{mildenhall2020nerf} represent complex scenes as Neural Radiance Fields (NeRF) for novel view synthesis by optimizing an implicit continuous volumetric scene function.
Due to the simplicity and extraordinary performance, NeRF~\cite{mildenhall2020nerf} has been extended to plenty of variants, \eg,  faster inference~\cite{reiser2021kilonerf,garbin2021fastnerf,rebain2021derf,rebain2021derf,lindell2021autoint}, pose estimation~\cite{YenChen20arxiv_iNeRF,lin2021barf,SCNeRF2021,meng2021gnerf,Wang21arxiv_nerfmm}, generalization~\cite{schwarz2020graf,chibane2021stereo,chen2021mvsnerf,yu2021pixelnerf,trevithick2020grf}, video~\cite{Xian20arxiv_stnif,Gao-freeviewvideo,Li20arxiv_nsff,li2021neural,peng2021neural}, and depth estimation~\cite{wei2021nerfingmvs}.

\noindent \textbf{3D-aware Image Synthesis.}
Several recent works have investigated how to  incorporate 3D representation into generative models~\cite{alhaija2018geometric,nguyen2019hologan,zhu2018visual,liao2020towards,nguyen2020blockgan,henderson2020leveraging,DeVries_2021_ICCV}.
Nguyen~\etal~\cite{nguyen2019hologan} combine a strong inductive bias about the 3D world with deep generative models to learn disentangled representations of 3D objects.
HoloGAN~\cite{nguyen2019hologan} provides control over the pose of generated objects through rigid-body transformations of the learned 3D features.
Schwarz~\etal~\cite{schwarz2020graf} propose GRAF, a generative model radiance fields model for 3D-aware image synthesis from unposed 2D images. 
pi-GAN~\cite{chan2021pi} adopts a SIREN-based neural implicit representation with periodic activation functions as the backbone of the generator.
By representing scenes as compositional generative neural feature fields, GIRAFFE~\cite{niemeyer2021giraffe} disentangles individual objects from the background. 
However, these methods optimize a single view of the generated scene independently and ignore the geometry constraints between views.

\section{Method} 
\label{sec:method}
Our goal is to generate photorealistic high-resolution images with explicit control over the camera pose while maintaining multi-view consistency.  
We now present the main components of the proposed method. 
First, we briefly review the background of  NeRF-based generative adversarial networks~\cite{schwarz2020graf,niemeyer2021giraffe,chan2021pi} and identify the limitations of previous methods~(see Sec.~\ref{sec:preliminaries}).
Second, we analyze the cause of the multi-view inconsistency problem and propose the image-level multi-view joint optimization to address this problem~(see Sec.~\ref{sec:image-level multi-view}). 
Besides, we design a two-stage training strategy that extends multi-view optimization to the feature level to generate high-resolution images with fine details.~(see Sec.~\ref{sec:feature-level}).
Finally, we describe the training details in Sec.~\ref{sec:training details}. 
Fig.~\ref{fig:framework} shows the framework of the proposed method.
\subsection{Preliminaries}
\label{sec:preliminaries}

\noindent \textbf{Neural Radiance Fields.}
Neural radiance field~(NeRF) synthesizes novel views of the scene by optimizing  a fully-connected network using a set of input views.
The MLP network maps a continuous 5D coordinate (3D location $\textbf{x}$ and 2D viewing direction $\textbf{d}$) to an emitted color $\textbf{c}$ and volume density $\sigma$~\cite{mildenhall2020nerf}:
\begin{equation}
\label{eq:nerf}
 (\gamma(\textbf{x}), \gamma(\textbf{d})) \xrightarrow[]{} (\textbf{c}, \sigma),
\end{equation}
where $\gamma$ indicates the positional encoding mapping function.
To render the neural radiance field from a viewpoint, Mildenhall~\etal~\cite{mildenhall2020nerf} use classic volume rendering to accumulate the output colors $\textbf{c}$ and densities $\sigma$ into an image.

\noindent \textbf{Generative Radiance Fields.}
Generative neural radiance fields aim to learn a model for synthesizing novel scenes by training on unposed 2D images.
Schwarz~\etal~\cite{schwarz2020graf} adopt an adversarial framework to train a generative model for radiance fields (GRAF).
The generative radiance field is conditioned on a shape code $z_s$ and an appearance code $z_a$:
\begin{equation}
\label{eq:graf}
(\gamma(\textbf{x}), \gamma(\textbf{d}), z_s, z_a)) \xrightarrow[]{} (\textbf{c}, \sigma).
\end{equation}
Following GRAF~\cite{schwarz2020graf}, Niemeyer~\etal~\cite{niemeyer2021giraffe} introduce a compositional generative neural feature field~(GIRAFFE).
Inspired by StyleGAN~\cite{karras2019style}, 
Chan~\etal~\cite{chan2021pi} instead propose periodic implicit generative adversarial networks~(pi-GAN) with feature-wise linear modulation (FiLM) conditioning.

\noindent\textbf{Limitations.}
We notice two limitations of existing approaches~\cite{schwarz2020graf,niemeyer2021giraffe,chan2021pi}.
First, they do not guarantee geometry constraints between different views.
Consequently, they usually suffer from collapsed results under large pose variations or have obvious inconsistent artifacts across views.
Second, the rendered high-resolution images typically lack realism and fine details due to the huge computational cost of NeRF model.

\subsection{Multi-view Joint Optimization}
\subsubsection{Image-level Multi-view Joint Optimization}
\label{sec:image-level multi-view}

\begin{figure}[]
\centering
  \includegraphics[width=0.8 \linewidth]{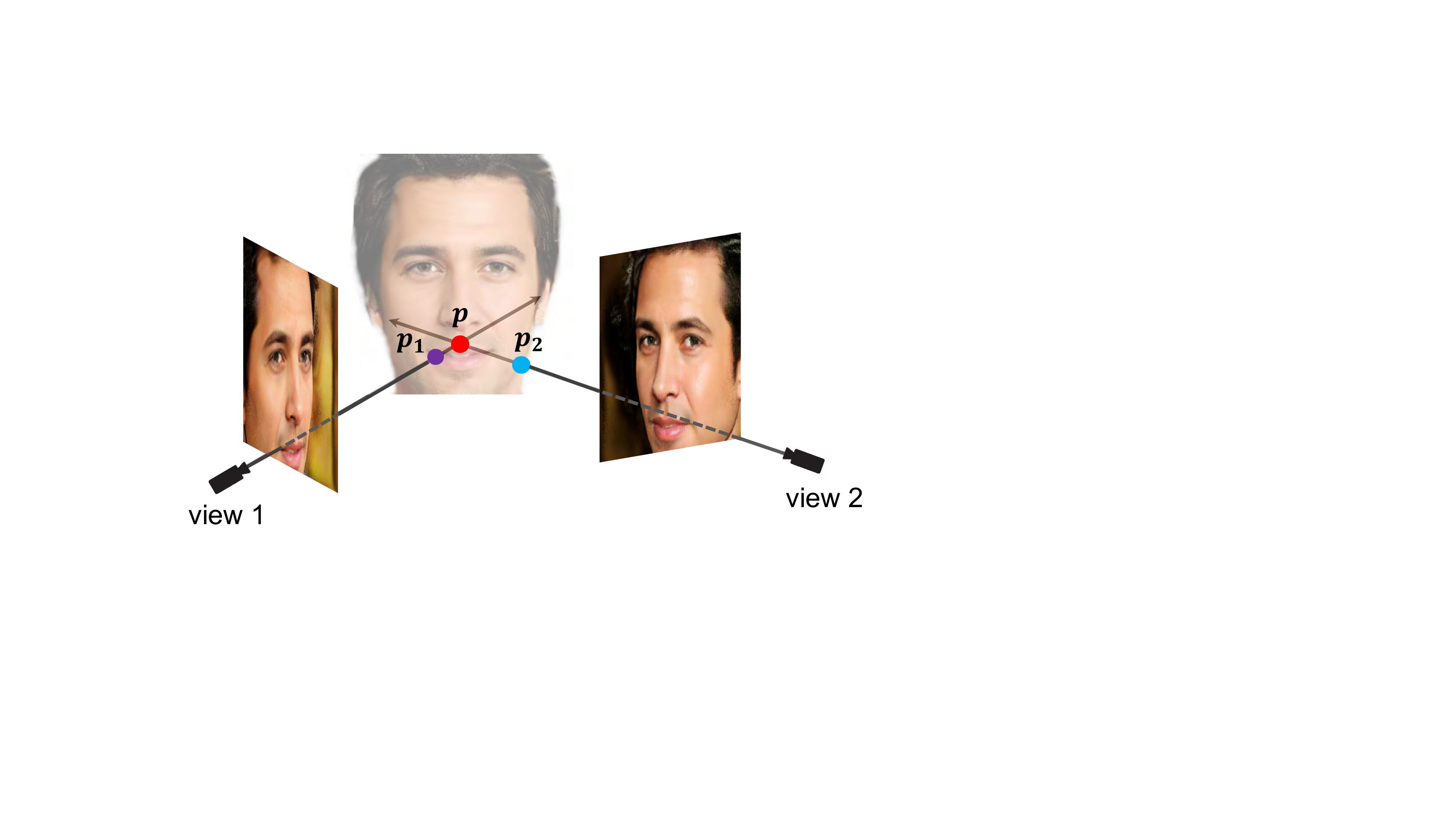}
  \vspace{-2mm} 
  \caption{\textbf{Visualization of shape-radiance ambiguity.} 
For illustration, we assume the $p$ (the \textcolor{red}{red} dot) is the location of correct geometry, and $p_1$ (the \textcolor{violet}{violet} dot) and $p_2$ (the \textcolor{blue}{blue} dot) are incorrect geometries.
In the absence of geometry constraints, the model can fit to incorrect geometry $p_1$ in view 1 and  $p_2$ in view 2 independently to simulate the effect of the correct geometry $p$. 
}
\label{fig:ambiguity}
\vspace{-3mm} 
\end{figure}
\begin{figure*}[]
\centering
  \includegraphics[width=1\linewidth]{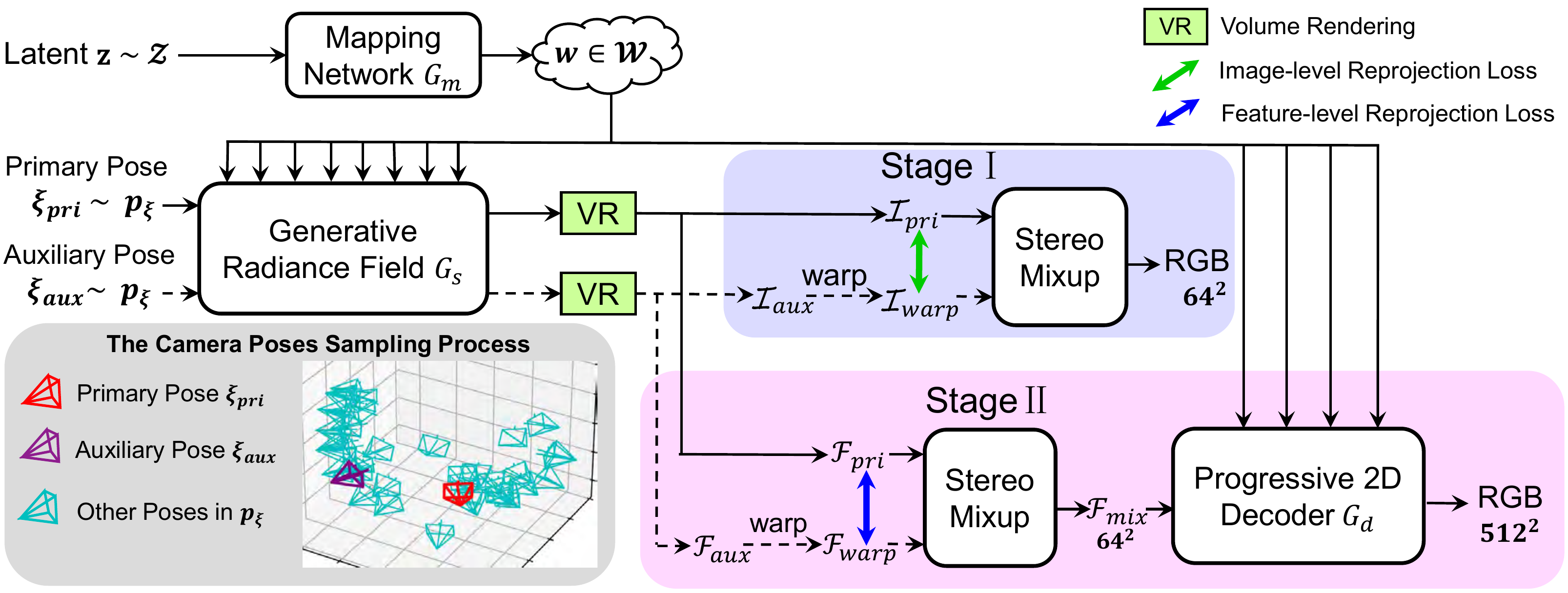}
  \vspace{-4mm}
  \caption{\textbf{The structure of generator $G_{\theta}$.}
During training, the generative radiance field network $G_s$ takes primary pose $\xi_{pri}$ and auxiliary pose $\xi_{aux}$ as input. 
The mapping network $G_m$ maps the input latent $z$ to intermediate latent $w$, which conditions both the  generative radiance field network $G_s$  and the progressive 2D decoder $G_d$.
In Stage~\uppercase\expandafter{\romannumeral1}, we directly render primary image $\mathcal{I}_{pri}$ and auxiliary image $\mathcal{I}_{aux}$ with the color and density output from $G_s$.
Then we perform image-level multi-view joint optimization and output a low-resolution
RGB image~($64^2$).
In Stage~\uppercase\expandafter{\romannumeral2}, we instead use volume rendering to accumulate 2D feature maps at low resolution~($64^2$), and then perform multi-view optimization at the feature level.
The progressive 2D decoder $G_d$ upsamples 2D feature map $\mathcal{F}_{mix}$ to a high-resolution RGB image~($128^2$, $256^2$, $512^2$) for fine 2D details.
During inference, only the primary pose is required without auxiliary pose~(the dotted lines do not participate in inference).
}
\vspace{-3mm}
\label{fig:framework}
\end{figure*}

\begin{figure}[]
\centering
  \includegraphics[width=0.95 \linewidth]{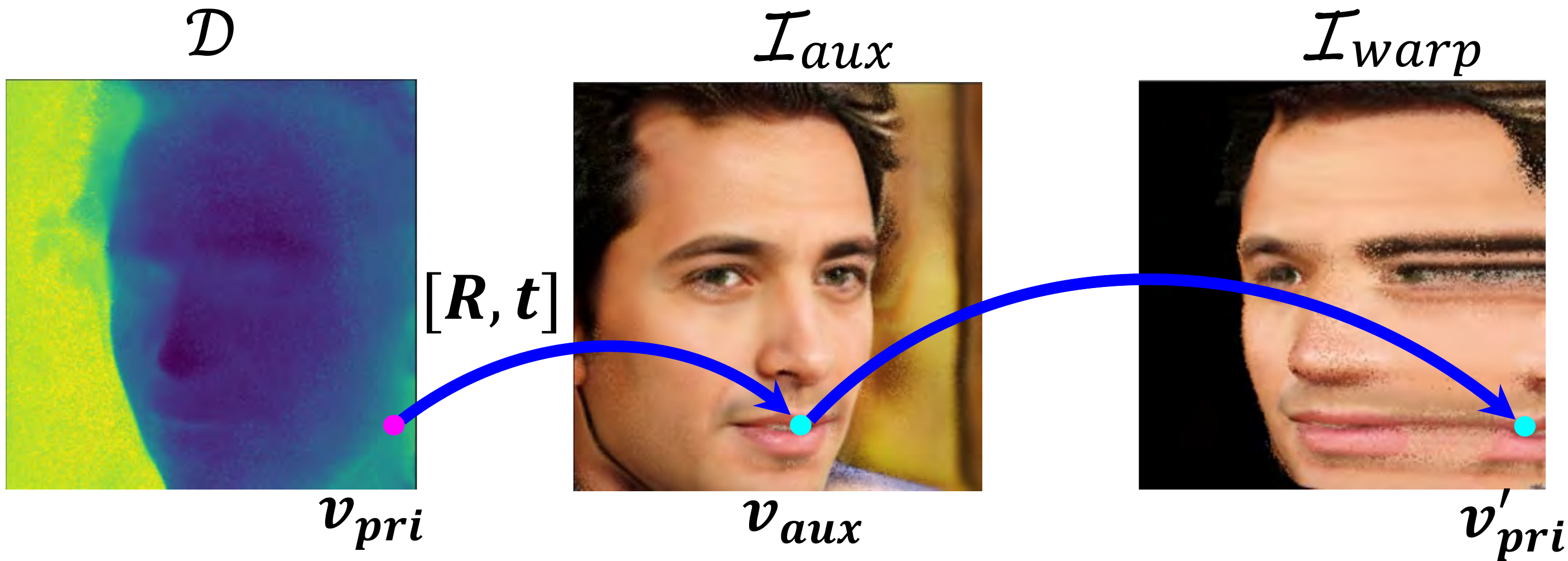}
  \vspace{-2mm} 
  \caption{\textbf{Illustration of the warping process.}
For each pixel $v_{pri}$ in the primary image  $\mathcal{I}_{pri}$, we first calculate the location of $v_{aux}$ (the corresponding pixel of $v_{pri}$ in the auxiliary image  $\mathcal{I}_{aux}$) based on the depth value $\mathcal{D}(v_{pri})$ and camera transformation matrix $[R, t]$.
Then we can reconstruct the pixel $v_{pri}'$ of the warped image $\mathcal{I}_{warp}$ from the primary view using the value of pixel~$v_{aux}$.
We observe that the warped image has a wrong appearance, which verifies the incorrect geometry shape learned by model. 
}
\vspace{-4mm}
\label{fig:warp}
\end{figure}

\noindent\textbf{Shape-radiance Ambiguity.} 
In this part, we analyze the cause of multi-view inconsistency problem in NeRF-based generative models.
We observe that optimizing the radiance fields from a set of 2D training images can encounter critical degenerate solutions in the absence of geometry constraints. 
This phenomenon is referred to as shape-radiance ambiguity~\cite{zhang2020nerf++}, in which the model can fit the training images with inaccurate 3D shape by a suitable choice of radiance field at each surface point~(see Fig~\ref{fig:ambiguity}).
To better illustrate the shape-radiance ambiguity~\cite{zhang2020nerf++}, we warp the rendered images from view 1 to view 2 based on the underlying depth and camera transformation matrix $[R, t]$~(see the details of warping process in Fig.~\ref{fig:warp} and Eq.~\ref{eq:transform}).
We find the warped image shows a wrong appearance, which verifies the assumption of degenerate solutions to the learned 3D shape.
To avoid the shape-radiance ambiguity~\cite{zhang2020nerf++}, NeRF~\cite{mildenhall2020nerf} requires a large number of posed training images from different input views for the scene.
However, generative radiance fields have neither annotated camera poses nor sufficient
multi-view images in the training dataset.
Consequently, the generative model can synthesize reasonable images in some views but produce poor renderings in other views~(see Fig.~\ref{fig:failture}).

\begin{figure}[]
\centering
  \includegraphics[width=0.95 \linewidth]{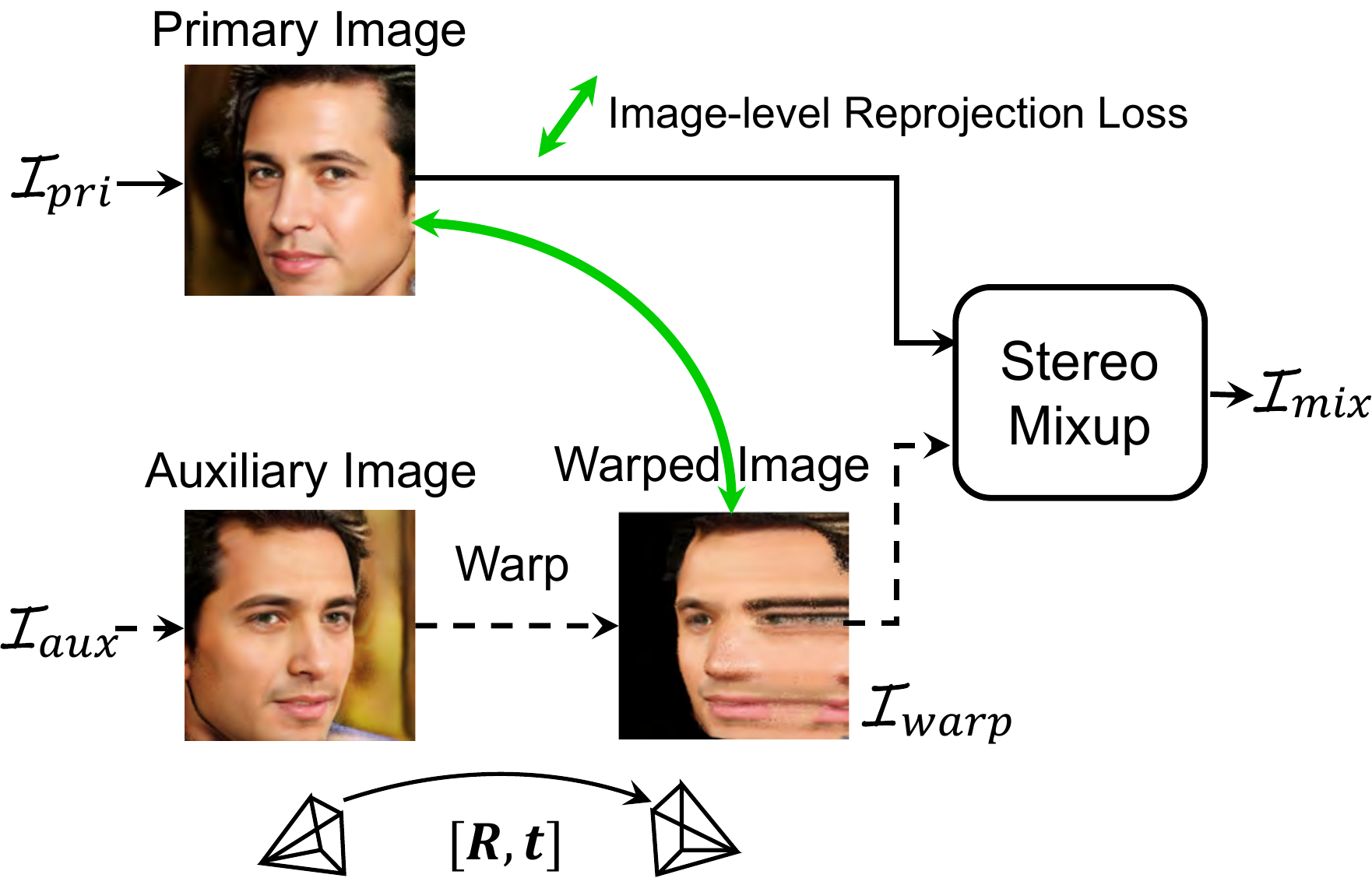} 
  \vspace{-1mm}
  \caption{\textbf{Image-level multi-view joint optimization.}
We enforce the photometric consistency between the primary image and the warped image by minimizing the image-level re-projection loss.
Besides, we integrate a stereo mixup module to encourage the warped image to be similar to a real image.
The dotted line does not participate in the inference stage.
}
\vspace{-2mm} 
\label{fig:multi_view}
\end{figure}

\noindent\textbf{Warping Process.} 
To alleviate the shape-radiance ambiguity~\cite{zhang2020nerf++},
we propose to establish multi-view geometry constraints~\cite{chen1993view,debevec1996modeling,andrew2001multiple,seitz1996view,zhou2017unsupervised,godard2019digging} via the warping process between views.
First, following pi-GAN~\cite{chan2021pi}, we adopt a style-based generator which contains a
synthesis network $G_s$~(a SIREN-based~\cite{sitzmann2020implicit,chan2021pi} generative radiance field) and a mapping network $G_m$~(a simple MLP network with ReLU)~(see Fig.~\ref{fig:framework}).
Given a latent code $z\in \mathbb{R}^{256}$ in the input latent space $\mathcal{Z}$, the mapping network $G_m$:$\mathcal{Z} \xrightarrow[ ]{}\mathcal{W} $ can produce the intermidiate latent  $w \in \mathbb{R}^{256}$, which controls the synthesis network $G_s$ at each layer.
Second, instead of only optimizing a single view independently, we aim to optimize multiple views jointly to maintain the 3D consistency across views.
As shown in the left of Fig.~\ref{fig:framework}, we randomly sample two camera poses, \ie,  the primary pose $\xi_{pri}$ and  the auxiliary pose $\xi_{aux}$, from the pose distribution $p_{\xi}$.
Taking $\xi_{pri}$ and $\xi_{aux}$ as input, the generative model $G_s$ synthesizes two views of the generated images separately: the primary image $\mathcal{I}_{pri}$ and the auxiliary image $\mathcal{I}_{aux}$.
Then we can build geometry constraints between $\xi_{pri}$ and $\xi_{aux}$
via image warping, which reconstructs the primary view by sampling pixels from the auxiliary image $\mathcal{I}_{aux}$.
Specifically, for each point $v_{pri}$ in the primary image $\mathcal{I}_{pri}$, we first find the corresponding pixel $v_{aux}$ in the auxiliary image $\mathcal{I}_{aux}$ through the stereo correspondence, and then reconstruct the pixel $v'_{pri}$  of the warped image $\mathcal{I}_{warp}$ in primary view using the value of $v_{aux}$~(see Fig.~\ref{fig:warp}).
Next, we present a detailed calculation procedure of the warping process.
The stereo correspondence is calculated based on the depth map $\mathcal{D}$ of the primary image and camera transformation matrix from  $\xi_{pri}$ to $\xi_{aux}$.
The depth can be rendered in a similar way as rendering the color image~\cite{mildenhall2020nerf,kangle2021dsnerf}.
Given the pixel $v_{pri}$ from the primary view, the depth value $\mathcal{D}(v_{pri})$ is formulated as:
\begin{equation}
\label{eq:depth}
\begin{split}
& \mathcal{D}(v_{pri}) = \sum\limits_{i=1}^{N} T_i(1 - exp(-\sigma_i\delta_i))d_i, \\
& \text{where~} T_i= exp(-\sum\limits_{j=1}^{i}\sigma_j\delta_j),
\end{split}
\end{equation}
where $N$ is the number of samples in the camera ray, $\delta_i = d_{i+1} - d_{i}$ is the distance between adjacent sample points and $\sigma_i$ is the volume density of sample $i$~(refer to~\cite{mildenhall2020nerf,kangle2021dsnerf} to see more details).
With the depth value $\mathcal{D}(v_{pri})$, we can obtain the homogeneous coordinates $h_{pri}$ of pixel $v_{pri}$ in the primary camera coordinate system through perspective projection.
Then the projected coordinates $h_{aux}$ in the auxiliary view can be calculated as:
\begin{equation}
\label{eq:transform}
 h_{aux} = K[R, t]\mathcal{D}(v_{pri})K^{-1}h_{pri},
\end{equation}
where the camera intrinsics $K$ are known parameters and
the camera transformation matrix $[R, t]$ can be calculated from the primary pose $\xi_{pri}$ and the auxiliary pose $\xi_{aux}$.
Finally, we can reconstruct the pixel $v'_{pri}$ in the warped image $\mathcal{I}_{warp}$ from the primary view using the value of pixel $v_{aux}$~(located in  $h_{aux}$ of $\mathcal{I}_{aux}$).

\noindent\textbf{Image-level Joint Optimization.}
After obtaining the warped image $\mathcal{I}_{warp}$, we perform image-level multi-view joint optimization by enforcing the photometric consistency and employing a stereo mixup module~(see Fig.~\ref{fig:multi_view}).
To satisfy the geometry constraints between views, we enforce the photometric consistency across views by minimizing the re-projection loss between the primary image $\mathcal{I}_{pri}$ and the warped image $\mathcal{I}_{warp}$.
Following the common practice in image reconstruction ~\cite{wang2004image,zhao2016loss,pillai2019superdepth,zhou2017unsupervised,godard2019digging,lyu2021hr}, we  formulate the image-level re-projection loss as the combination of L1~\cite{zhao2016loss} and SSIM~\cite{wang2004image}:
\begin{equation}
\label{eq:re}
L_{ir} = (1 - \mu)||\mathcal{I}_{pri} - \mathcal{I}_{warp}||_1 + \frac{\mu}{2} (1 - SSIM(\mathcal{I}_{pri}, \mathcal{I}_{warp})),
\end{equation}
where SSIM is a perceptual metric of image structural similarity and $\mu=0.85$ empirically.
In addition to being similar to the primary image, the warped image should also look like a real image.
A straightforward method is introducing two discriminators. 
One is to compare the warped image $\mathcal{I}_{warp}$  with an arbitrary real image sampled from the training dataset, and the other one compares the primary image $\mathcal{I}_{pri}$.
However, introducing extra modules can increase the computation complexity.
Inspired by the \textit{mixup} strategy~\cite{zhang2018mixup}, we instead propose a stereo mixup module to optimize both $\mathcal{I}_{pri}$ and $\mathcal{I}_{warp}$ by constructing a virtual mixed image:
\begin{equation}
\label{eq:mixup}
\mathcal{I}_{mix} = \eta\mathcal{I}_{pri} + (1 - \eta)\mathcal{I}_{warp},
\end{equation}
where $\eta$ is a dynamic number randomly sampled from the range of $[0,1]$ in every training iteration, and
$\mathcal{I}_{mix}$ is the input of discriminator.
It is worth noting that the auxiliary pose is introduced to construct the geometry constraints, and thus only required in the training process.
In the inference stage, the generative model only takes the primary pose $\xi_{pri}$ and latent code $z$ as input to generate the primary image directly.

\begin{figure}[]
\centering
  \includegraphics[width=0.8\linewidth]{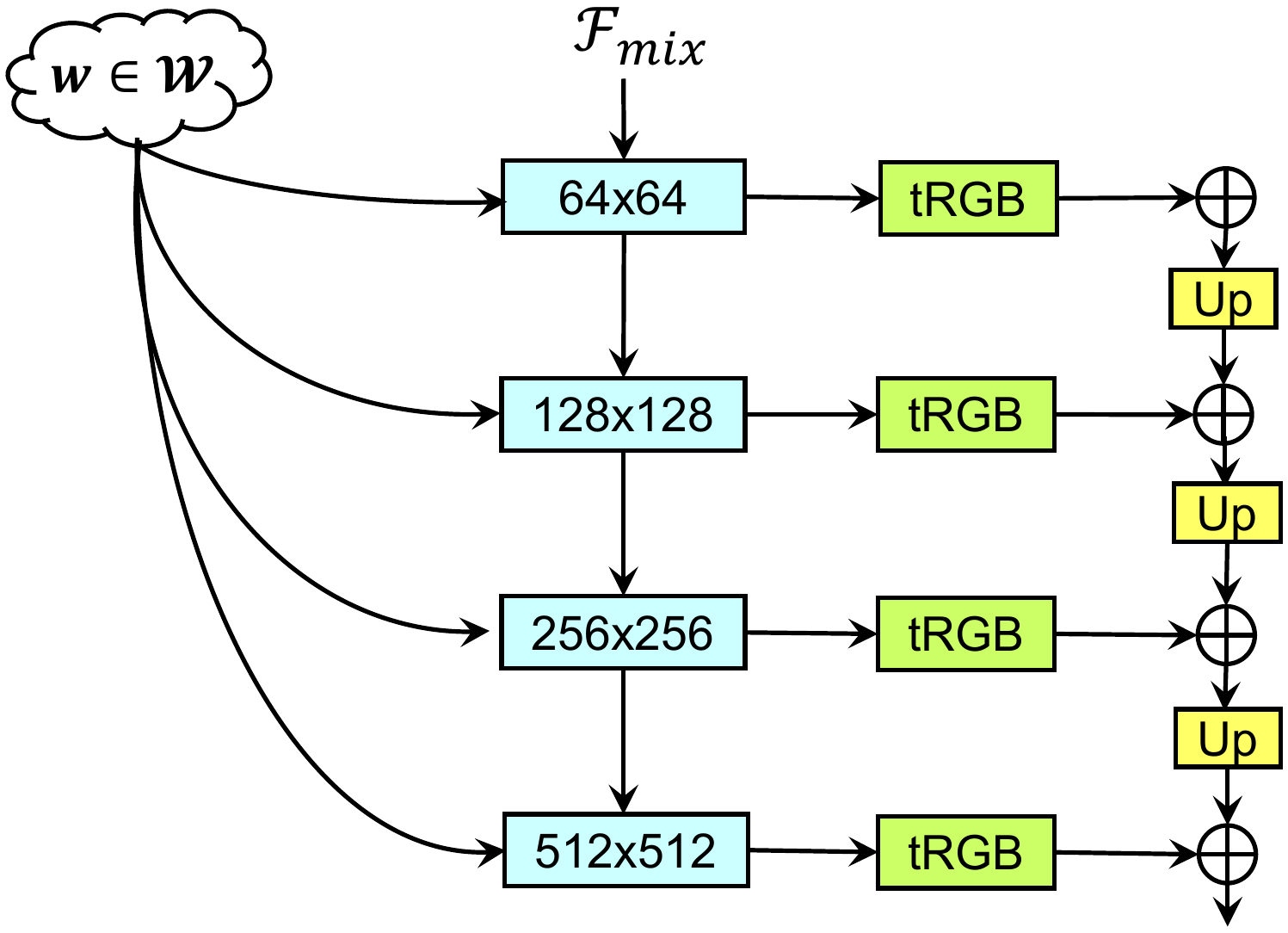}
  \vspace{-2mm}
  \caption{\textbf{Progressive 2D decoder $G_d$.}
During training, the decoder takes the stereo mixup feature $\mathcal{F}_{mix}$ (produced by $\mathcal{F}_{pri}$ and $\mathcal{F}_{warp}$)  as input at low resolution~($64^2$).
Then the intermediate latent $w$ conditions the decoder at each layer.
Here \boxed{\text{tRGB}} denotes the 1x1 convolutions which convert the high-dimensional features to RGB images, and \boxed{\text{Up}} denotes the bilinear upsampling operation.
}
\label{fig:decoder}
\vspace{-2mm}
\end{figure}

\subsubsection{Feature-level Multi-view Joint Optimization}
\label{sec:feature-level}
In practice, we also encounter one practical challenge: NeRF-based generative models~\cite{schwarz2020graf,niemeyer2021giraffe,chan2021pi} typically struggle to render high-resolution images with fine details due to the huge computational of NeRF~\cite{mildenhall2020nerf} model.
To render images with both fine 2D details and correct 3D shape, we design a two-stage training strategy and extend multi-view optimization to the feature level.  
We begin training at a low resolution ($64^2$) in Stage \uppercase\expandafter{\romannumeral1}, and then increase to high resolutions~($128^2$, $256^2$, $512^2$) in Stage \uppercase\expandafter{\romannumeral2}~(see Fig.~\ref{fig:framework}).
In Stage \uppercase\expandafter{\romannumeral1}, we directly render primary and auxiliary images with the color and density output from the generative radiance field network $G_s$.
With the guidance of geometry constraints, we perform image-level multi-view joint optimization to enhance the geometric reasoning ability of the model. 
In Stage \uppercase\expandafter{\romannumeral2}, to alleviate the computation-intensive problem of rendering high-resolution images, we instead train the model via feature-level multi-view optimization for better visual quality. 
First, we adopt a hybrid MLP-CNN architecture to disentangle the geometry of 3D shape from fine details of 2D appearance.
Then we generalize volume rendering~\cite{niemeyer2021giraffe} to the feature level by rendering 2D primary feature map $\mathcal{F}_{mix}$ at low resolution~($64^2$):
\begin{equation}
\label{eq:feaure}
\begin{split}
& \mathcal{F}_{pri} = \sum\limits_{i=1}^{N} T_i(1 - exp(-\sigma_i\delta_i))f_i,
\end{split}
\end{equation}
where $f_i \in \mathbb{R}^{256}$ is the feature before the final layer of $G_s$, and other symbols are defined in Eq.~\ref{eq:depth}.
The auxiliary feature map $\mathcal{F}_{aux}$ is rendered in the same way as $\mathcal{F}_{pri}$, and the warped feature map $\mathcal{F}_{warp}$ can be obtained through the warping process.
Second, we perform multi-view feature-level joint optimization on low-resolution feature maps~($64^2$).
To enforce the geometry consistency in the feature space,  we take the implicit diversified Markov Random Fields~(MRF) loss~\cite{wang2018image} as the feature-level re-projection loss:
\begin{equation}
\label{eq:mrf}
L_{fr} = L_{mrf}(\mathcal{F}_{pri}, \mathcal{F}_{warp}),
\end{equation}
which can encourage the model to capture high-frequency geometry details~\cite{feng2021learning}.
Then the stereo mixup mechanism is also applied to the 2D feature maps: $\mathcal{F}_{mix} = \eta\mathcal{F}_{pri} + (1 - \eta)\mathcal{F}_{warp}$.
Third, we increase the resolution with a style-based 2D decoder~\cite{karras2019style} $G_d$, which takes $\mathcal{F}_{mix}$ as input and then upsamples to high-resolution RGB image~(see Fig.~\ref{fig:decoder}).
The 2D decoder $G_d$  is conditioned by the mapping network $G_m$ through adaptive instance normalization~(AdaIN)~\cite{huang2017arbitrary,dumoulin2016learned,karras2019style}.
As training progresses, we adopt the progressive growing strategy to grow the generator for higher resolution~\cite{karras2018progressive}.
When new layers are added to $G_d$, we use skip connections to fade the inserted layers in smoothly to stabilize and speed up the training process~\cite{karras2018progressive,karras2020analyzing}.

\subsection{Training Details}
\label{sec:training details} 
We use a progressive growing convolutional discriminator $D_{\phi}$ to compare the fake image produced by generator $G_{\theta}$ and real image $\mathcal{I}$ sampled from the training data with distribution $p_\mathcal{D}$.
We train MVCGAN using a non-saturating GAN objective with $R_1$ gradient penalty~\cite{mescheder2018training} and the proposed geometry-constrained objective $L_{re}$ as the total loss:
\begin{equation}
\label{eq:stage1}
\begin{split}
& \mathcal{V}(\theta, \phi) = \textbf{E}_{z\sim\mathcal{Z}, \xi_{pri}\sim p_{\xi}, \xi_{aux}\sim p_{\xi}}[f(D_{\phi}(G_{\theta}(z, \xi_{pri}, \xi_{aux}))] \\
& + \textbf{E}_{\mathcal{I}\sim p_{\mathcal{D}}}[f(-D_{\phi}(\mathcal{I})) -\lambda|| \nabla  D_{\phi}(\mathcal{I})||^2] + L_{re}, \\
\end{split}
\end{equation}
where $f(t)=-log(1+exp(-t))$,  $L_{re} = L_{ir}$ for Stage \uppercase\expandafter{\romannumeral1}~(see Eq.~\ref{eq:re}),  $L_{re} = L_{fr}$ for Stage \uppercase\expandafter{\romannumeral2}~(see Eq.~\ref{eq:mrf}), and $\lambda=10$.
More implementation details can be found in the supplementary material.

\begin{figure*}[]
\centering
\subfloat[Results on FFHQ~\cite{karras2019style}.]{%
  \includegraphics[width=0.96 \linewidth]{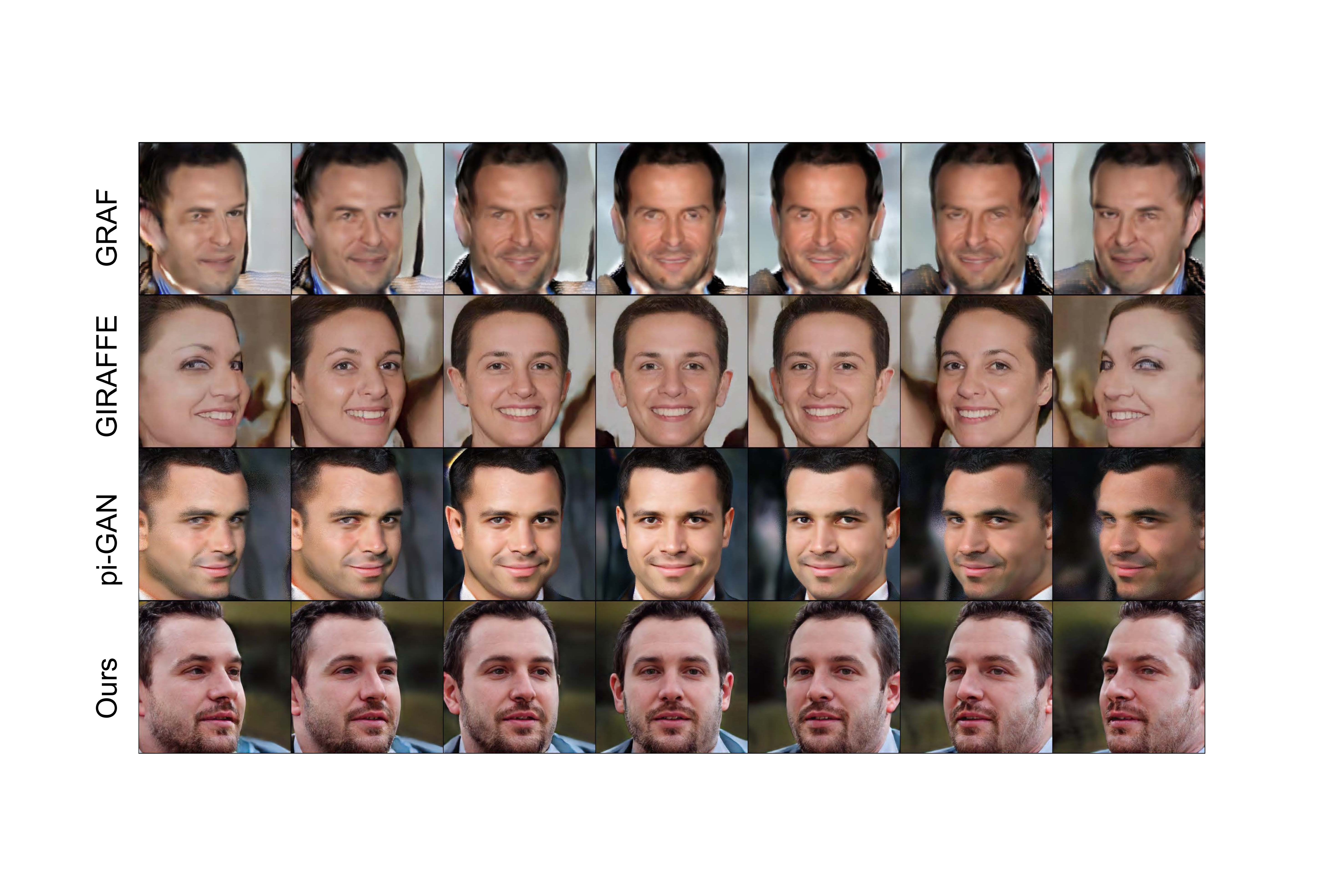}
}
\vspace{1mm}
\subfloat[Results on AFHQv2~\cite{choi2020stargan}.]{%
  \includegraphics[width=0.96\linewidth]{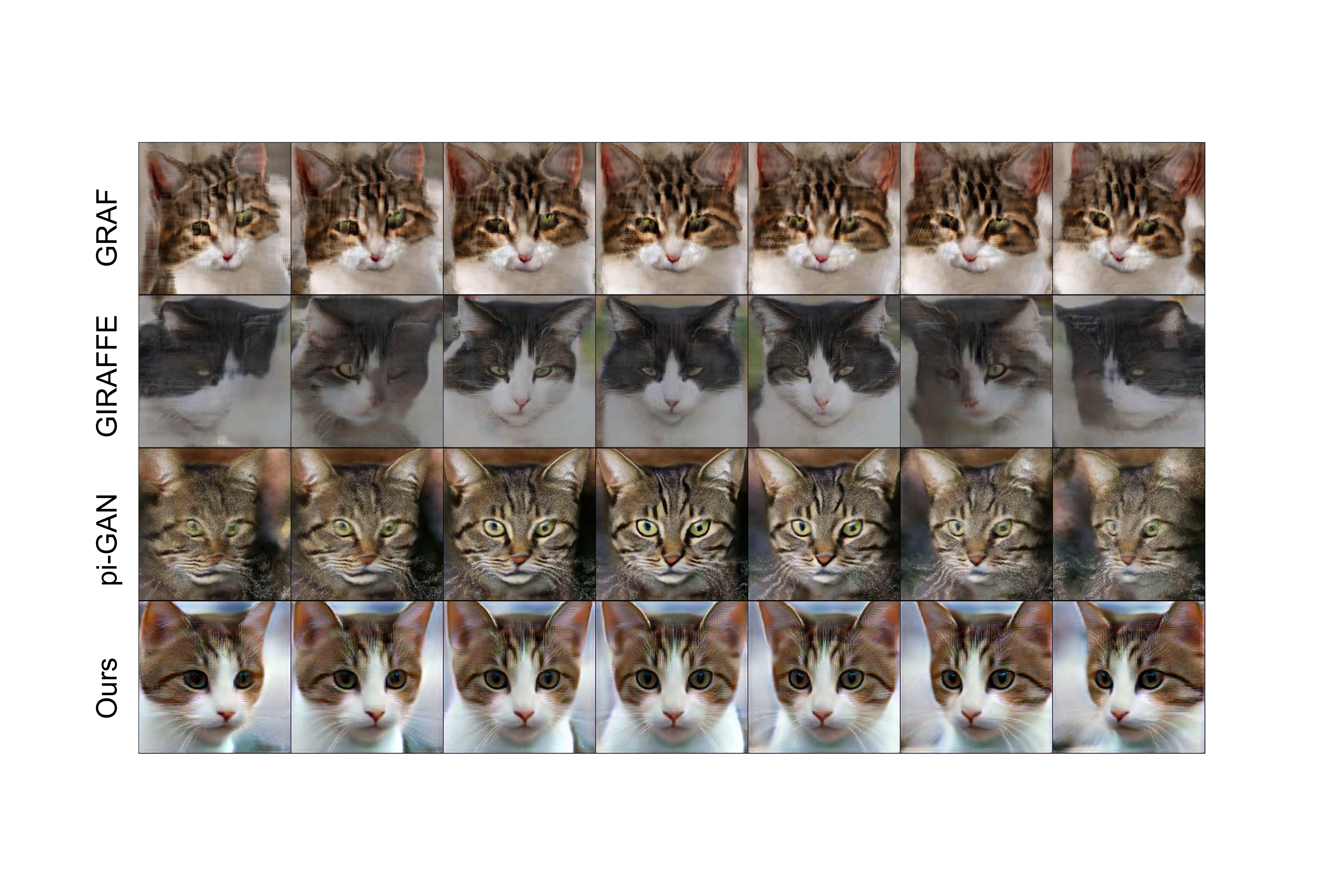}
}
\vspace{-3mm}
\caption{Qualitative comparison at $512^2$ resolution with GRAF~\cite{schwarz2020graf}, GIRAFFE~\cite{niemeyer2021giraffe}, and pi-GAN~\cite{chan2021pi}.}
\label{fig:comp}
\vspace{-4mm}
\end{figure*}

\section{Experiments}
\label{sec:exp}
\subsection{Experimental Settings}
\noindent\textbf{Datasets.}
We conduct experiments on three widely-used high-resolution image datasets: CELEBA-HQ~\cite{karras2018progressive}, FFHQ~\cite{karras2019style}, and AFHQv2~\cite{choi2020stargan}.
We choose the cat face images in the AFHQv2~\cite{choi2020stargan} dataset to conduct experiments for a fair comparison with previous works~\cite{schwarz2020graf,niemeyer2021giraffe,chan2021pi}.

\begin{table}
\centering
{
\setlength{\tabcolsep}{1.5mm}
\begin{tabular}{l|cc|cc|c}
\Xhline{2\arrayrulewidth}
 &  \multicolumn{2}{c|}{CELEBA-HQ} & \multicolumn{2}{c|}{FFHQ} & AFHQv2 \\
&   $256^2$ & $512^2$ & $256^2$ & $512^2$ &  $256^2$  \\
\hline
GRAF~\cite{schwarz2020graf} &  47.5 & 57.7 & 67.2 & 71.2 & 75.8  \\
pi-GAN~\cite{chan2021pi} & 39.7  & 41.8 & 38.1 & 39.9&  42.0  \\
GIRAFFE~\cite{niemeyer2021giraffe} & 36.0 & 36.2 & 34.6  & 37.7 & 29.2   \\
Ours & \textbf{11.8} & \textbf{12.9} & \textbf{13.7} & \textbf{13.4} & \textbf{17.1} \\
\Xhline{2\arrayrulewidth}
\end{tabular}}
\vspace{-2mm}
\caption{Quantitative comparison. We calculate FID between 20,000 generated and real images.} 
\vspace{-2mm}
\label{table:comp}
\end{table}

\subsection{Comparison with SOTA}
For quantitative comparison, we report Frechet Inception Distance (FID)~\cite{heusel2017gans} to evaluate image quality.
We compare our approach against three state-of-the-art 3D-aware image synthesis methods: GRAF~\cite{schwarz2020graf}, GIRAFFE~\cite{niemeyer2021giraffe} and pi-GAN~\cite{chan2021pi}.
As shown in Tab.~\ref{table:comp}, our method consistently outperforms other methods~\cite{schwarz2020graf,niemeyer2021giraffe,chan2021pi} on all datasets~\cite{karras2018progressive,karras2019style,choi2020stargan} by a large margin.
We also visualize the generated images on FFHQ~\cite{karras2019style} and AFHQv2~\cite{choi2020stargan} datasets for qualitative comparison.
As illustrated in Fig.~\ref{fig:comp}, we render images from a wide range of viewpoints.
We observe that GRAF~\cite{schwarz2020graf}, GIRAFFE~\cite{niemeyer2021giraffe} and pi-GAN~\cite{chan2021pi} either fail to synthesize reasonable results under large view variations or have obvious multi-view inconsistent artifacts.
By comparison, our method achieves the best performance both in visual quality and multi-view consistency.
Please refer to the supplementary material for more visualization results.
\begin{figure*}[]
\centering
  \includegraphics[width=1 \linewidth]{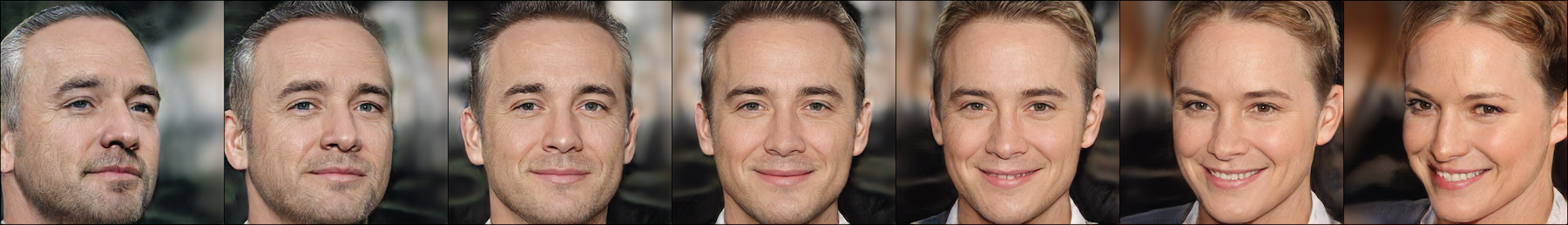}
  \vspace{-6mm}
  \caption{\textbf{Style interpolation.} We perform linear interpolation both in the intermediate latent  and camera pose space.
} 
\label{fig:style_inter}
\vspace{-4mm}
\end{figure*}

\begin{figure}[]
\centering

\subfloat[With image-level multi-view join optimization~(FID=22.5).]{%
  \includegraphics[width=1\linewidth]{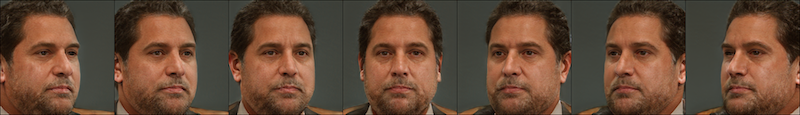}
}
\vspace{1mm}
\subfloat[With feature-level multi-view join optimization~(FID=13.7).]{%
  \includegraphics[width=1\linewidth]{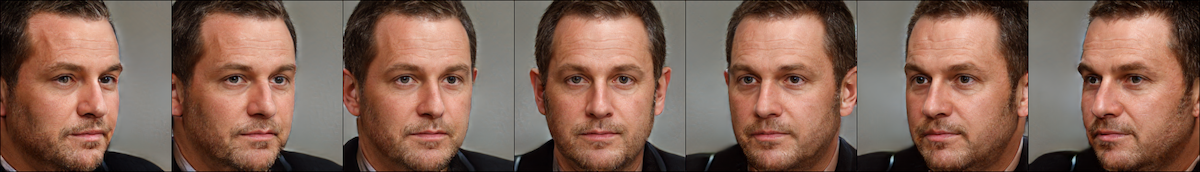}
}
\vspace{-2mm}
\caption{Ablation study on FFHQ~\cite{karras2019style} at $256^2$ resolution.}
\label{fig:abl}
\vspace{-4mm}
\end{figure}
\subsection{Ablation Studies}
\noindent\textbf{Image-level and Feature-level Optimization.}
We conduct ablation studies to help understand the individual contributions of image-level and feature-level multi-view joint optimization.
From Fig.~\ref{fig:abl}~(a), we observe that the generated images maintain the multi-view consistency under poses variations~(FID=22.5), indicating the image-level optimization can guide the model to learn a reasonable 3D shape.
With feature-level optimization~(see Fig.~\ref{fig:abl}~(b)), our approach can further improve the visual quality of generated images with fine 2D details~(FID=13.7).

\begin{figure}[]
\centering
  \includegraphics[width=1\linewidth]{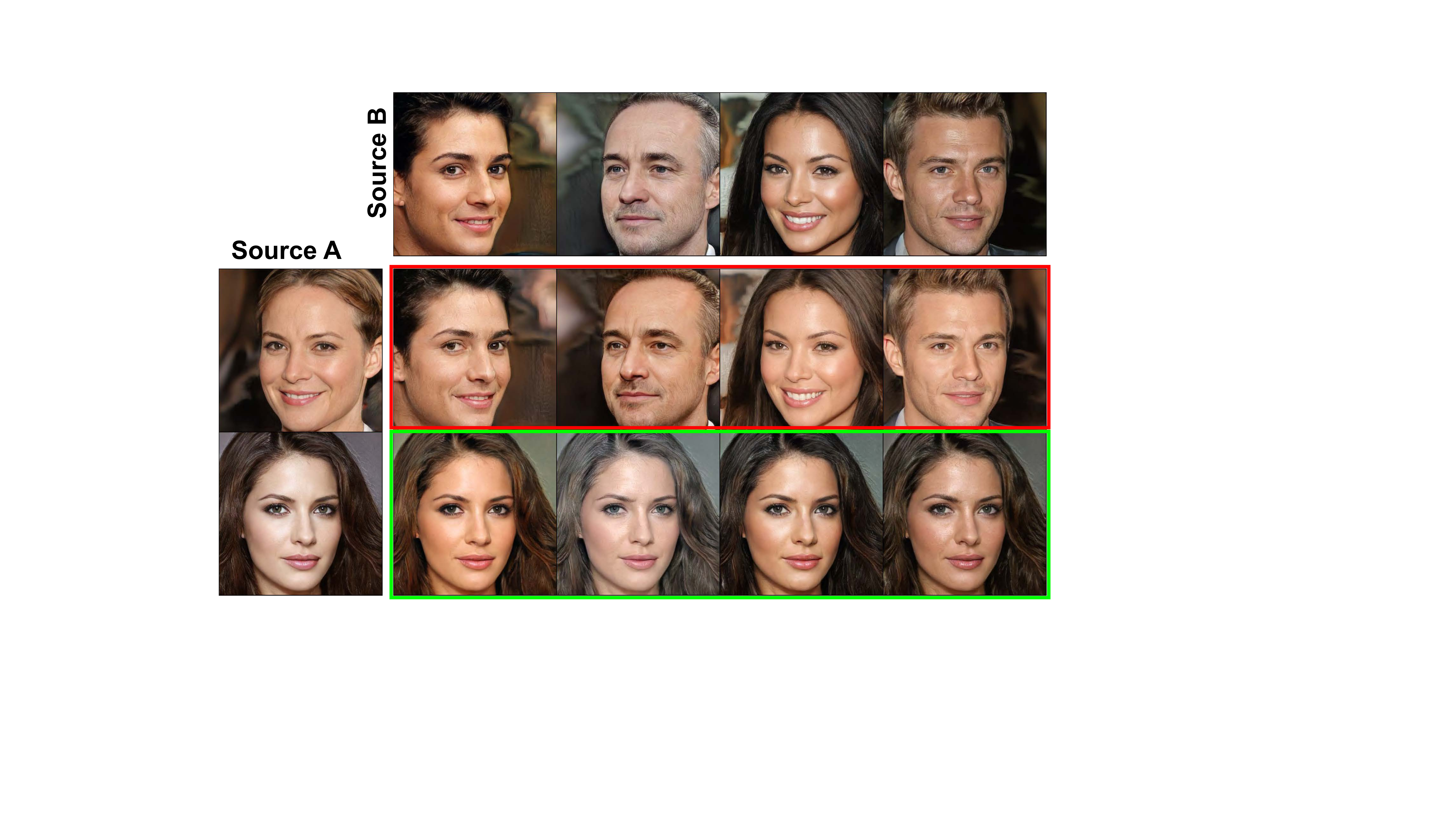}
  \vspace{-5mm}
  \caption{\textbf{Style mixing.}
  The source A and B images are generated from their input latent codes $z_A$ and $z_B$.
  The images in the  \textcolor{red}{red} box are generated by applying the $w_B$~(the intermediate latent corresponding to $z_B$) to $G_s$ and $w_A$~(corresponding to $z_A$) to $G_d$.
 The images in the \textcolor{green}{green} box are generated by applying the $w_A$ to $G_s$ and $w_B$ to $G_d$.
}
\label{fig:style_mixing}
\vspace{-2mm}
\end{figure}

\noindent\textbf{Shape-detail Disentanglement.}
Besides, we design a style mixing experiment to study what kinds of representations the generative radiance field $G_s$ and progressive 2D decoder $G_d$ learned respectively.
Specifically, we input two latent codes $z_A$ and $z_B$ into the mapping network $G_m$, and obtain the corresponding intermediate latent $w_A$, $w_B$ in $\mathcal{W}$ space.
Then we can generate style mixing images by applying $w_A$ and $w_B$ to control the different parts of the generator~($G_s$ and $G_d$).
As shown in Fig.~\ref{fig:style_mixing}, we observe that controlling $G_s$ changes the 3D shape~(identity and pose) while controlling $G_d$ changes 2D appearance details~(colors of skins, hair, and beard).
The results verify that the hybrid MLP-CNN architecture can disentangle the geometry of 3D shape from fine details of 2D appearance.

\noindent\textbf{Style Interpolation.}
We also conduct the style interpolation experiments to investigate the intermediate latent $w$ learned by the mapping network $G_m$.
Given two generated images, we perform linear interpolation both in the intermediate latent space $\mathcal{W}$ and the camera pose space.
As illustrated in Fig.~\ref{fig:style_inter}, the smooth transition of both pose and appearance demonstrates that our model learns semantically meaningful intermediate latent space $\mathcal{W}$.
\begin{figure}[]
\centering
  \includegraphics[width=1\linewidth]{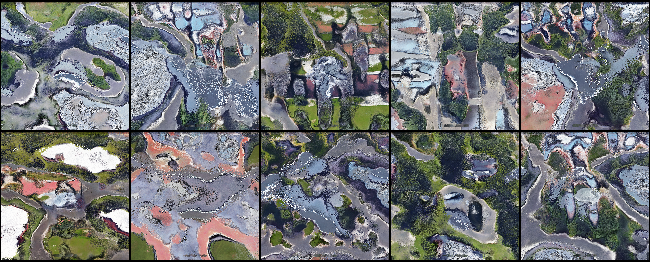}
  \vspace{-5mm}
  \caption{\textbf{Failure cases.}
   Our method does not perform well in scenarios with multiple objects and complex backgrounds. 
   For example, our model fails to synthesize high-quality images on the University-1652 dataset~\cite{zheng2020university}.}
\label{fig:failure_case}
\vspace{-5mm}
\end{figure}

\section{Conclusion and Discussion}
\label{sec:conclusion}
We present a multi-view consistent generative model (MVCGAN) for 3D-aware image synthesis.
The key idea underpinning the proposed method is to enhance the geometric reasoning ability of the generative model by introducing geometry constraints.
Extensive experiments demonstrate that MVCGAN achieves the state-of-the-art performance for 3D-aware image synthesis. 

\noindent\textbf{Limitations and future work.}
In this paper, our method mainly focuses on single-object scenes with simple backgrounds, and does not work well in multi-object and complex background-attached scenes~(see Fig.~\ref{fig:failure_case}).
To extend to the scenarios with complex background and multiple objects, one possible way is to learn a compositional radiance field that can model the foreground and background separately~\cite{yang2021learning}.
To render the whole scene, the geometry relationships between foreground objects and the background can be established by combing depth maps and occlusion maps.
In the future, we will incorporate extra image annotations to handle more complex real-world scenarios.

\clearpage
{\small
\bibliographystyle{ieee_fullname}
\bibliography{egbib}
}

\clearpage

\appendix

\section*{\Large\textbf{Appendix}}
\textit{
In the supplementary document, we first present the implementation details in Sec.~\ref{sec:imp}.
Next, we provide additional visualization results in Sec.~\ref{sec:add}.}

\section{Implementation Details}
\label{sec:imp}
In this section, we first present the network architectures of the generative radiance field $G_s$, the mapping network $G_m$,  the progressive 2D decoder $G_d$, and the discriminator $D_{\phi}$ in Sec.~\ref{sec:arch}.
Second, we discuss the training protocol in Sec.~\ref{sec:pro}.
Third, 
we describe the datasets used in experiments~(see Sec.~\ref{sec:data}).
Finally, 
we provide the details of compared methods in Sec.~\ref{sec:base}.

\subsection{Network Architectures}
\label{sec:arch}
\noindent \textbf{Generative Radiance Field.}
The generative radiance field network $G_s$ is a 8-layer SIREN-based MLP with periodic activation functions~\cite{sitzmann2020implicit}.
The dimension of the hidden layers is 256.

\noindent \textbf{Mapping Network.}
The mapping network $G_m$ is a 4-layer MLP network with leakyReLU as the activation function.
The dimension of the hidden layers is 256.
We sample the input latent code $z$ from a 256-dimensional standard Gaussian distribution. 

\noindent \textbf{Progressive 2D Decoder.}
The progressive 2D decoder $G_d$ is a fully-convolution neural network, which decreases the feature dimension from 256~(at $64^2$) to 32~(at $512^2$).

\noindent \textbf{Discriminator.}
The discriminator $D_{\phi}$ is a progressive growing convolutional network, which uses eight layers for $64^2$ and fourteen layers for $512^2$.

\begin{figure*}[]
\centering
  \includegraphics[width=1.0 \linewidth]{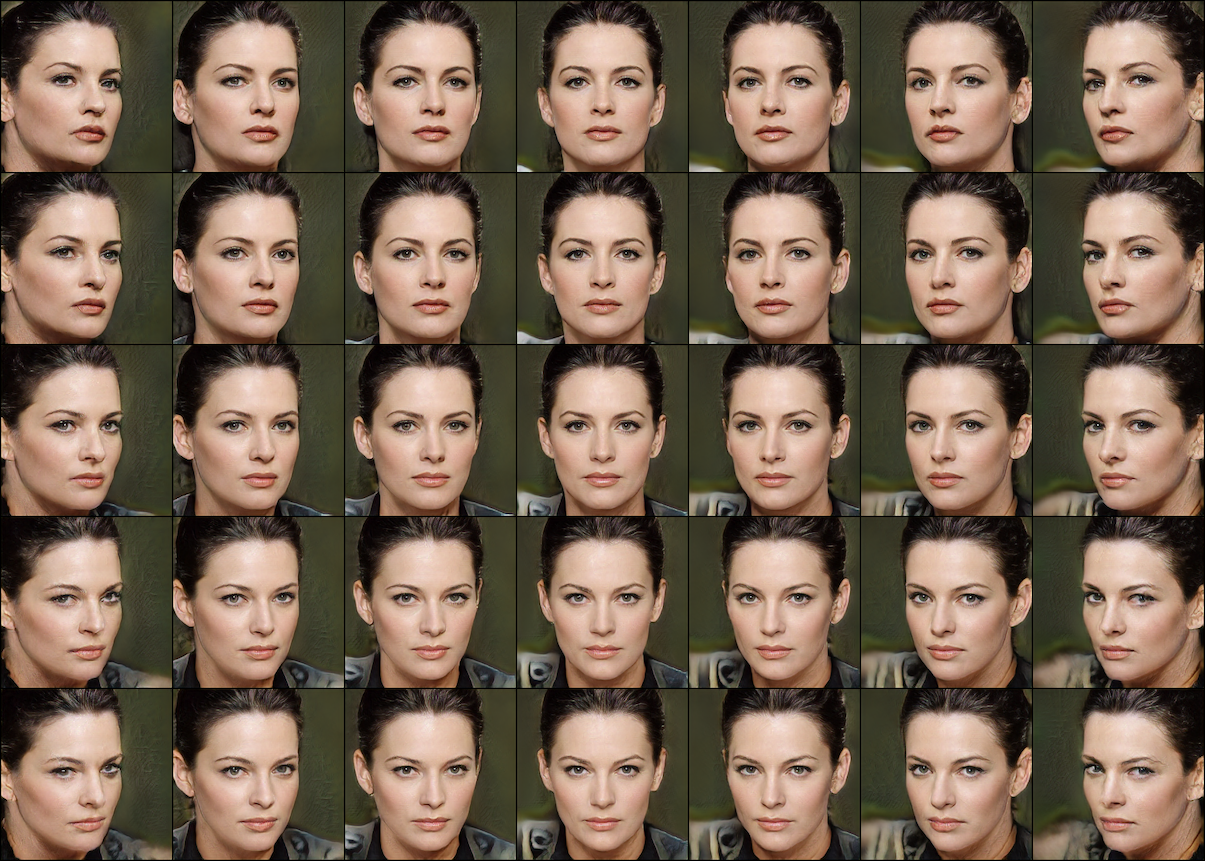}
  \caption{The images are rendered from 35 camera poses at resolution $256^2$ .
}
\label{fig:full_view}
\end{figure*}

\subsection{Training Protocol}
\label{sec:pro}
We employ Adam optimizer~\cite{kingma2014adam} with $\beta_1 = 0$, $\beta_2 = 0.9$, and the batch size of 56 for optimization.
The initial learning rate is set to $6.0\times10^{-5}$ for the generator and $2.0\times10^{-4}$ for the discriminator, and decay over training to $1.5\times10^{-5}$ and $5.0\times5^{-5}$ respectively.
We use 12 samples per ray for all datasets without hierarchical sampling strategy~\cite{mildenhall2020nerf,chan2021pi}.

\begin{figure*}
{
      \includegraphics[width=0.18\linewidth]{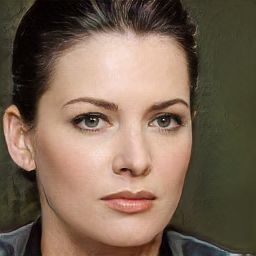}
     \hspace{0.1mm}
      \includegraphics[width=0.18\linewidth]{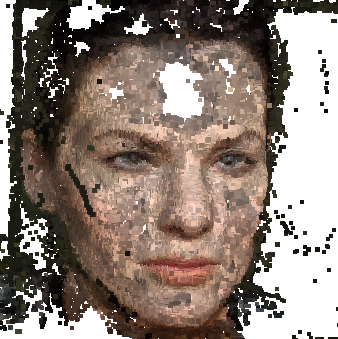}
     \hspace{0.05mm}
    \includegraphics[width=0.18 \linewidth]{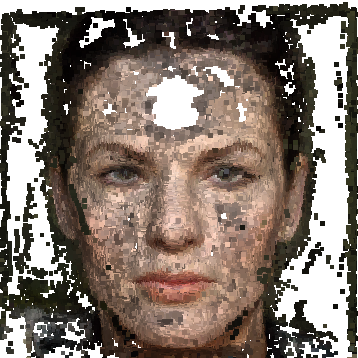}
    \hspace{0.05mm}
      \includegraphics[width=0.18 \linewidth]{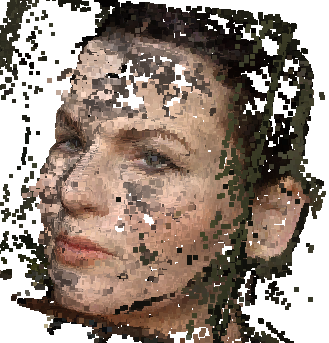}
     \hspace{0.05mm}
      \includegraphics[width=0.18 \linewidth]{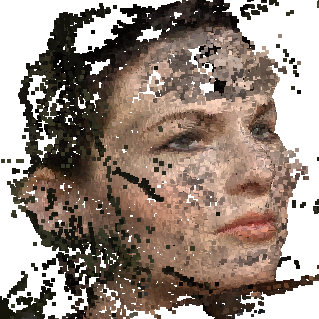}
      \includegraphics[width=0.18\linewidth]{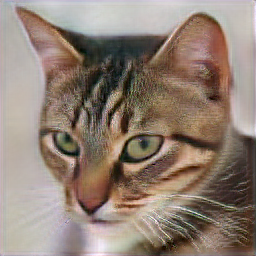}
     \hspace{2.5mm}
      \includegraphics[width=0.18\linewidth]{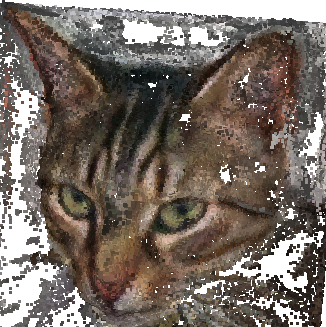}
     \hspace{2mm}
      \includegraphics[width=0.18 \linewidth]{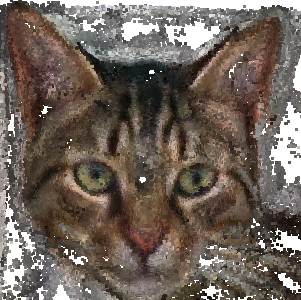}
     \hspace{2mm}
     \includegraphics[width=0.18 \linewidth]{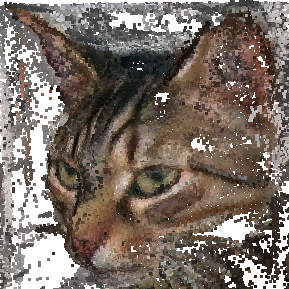}
     \hspace{2mm}
      \includegraphics[width=0.18 \linewidth]{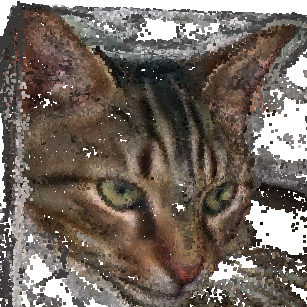}
}
  \caption{COLMAP reconstruction~\cite{schonberger2016structure} from synthesized images at resolution $256^2$.
  }
\label{fig:colmap}
\end{figure*}

\begin{figure*}[]
\centering
  \includegraphics[width=1 \linewidth]{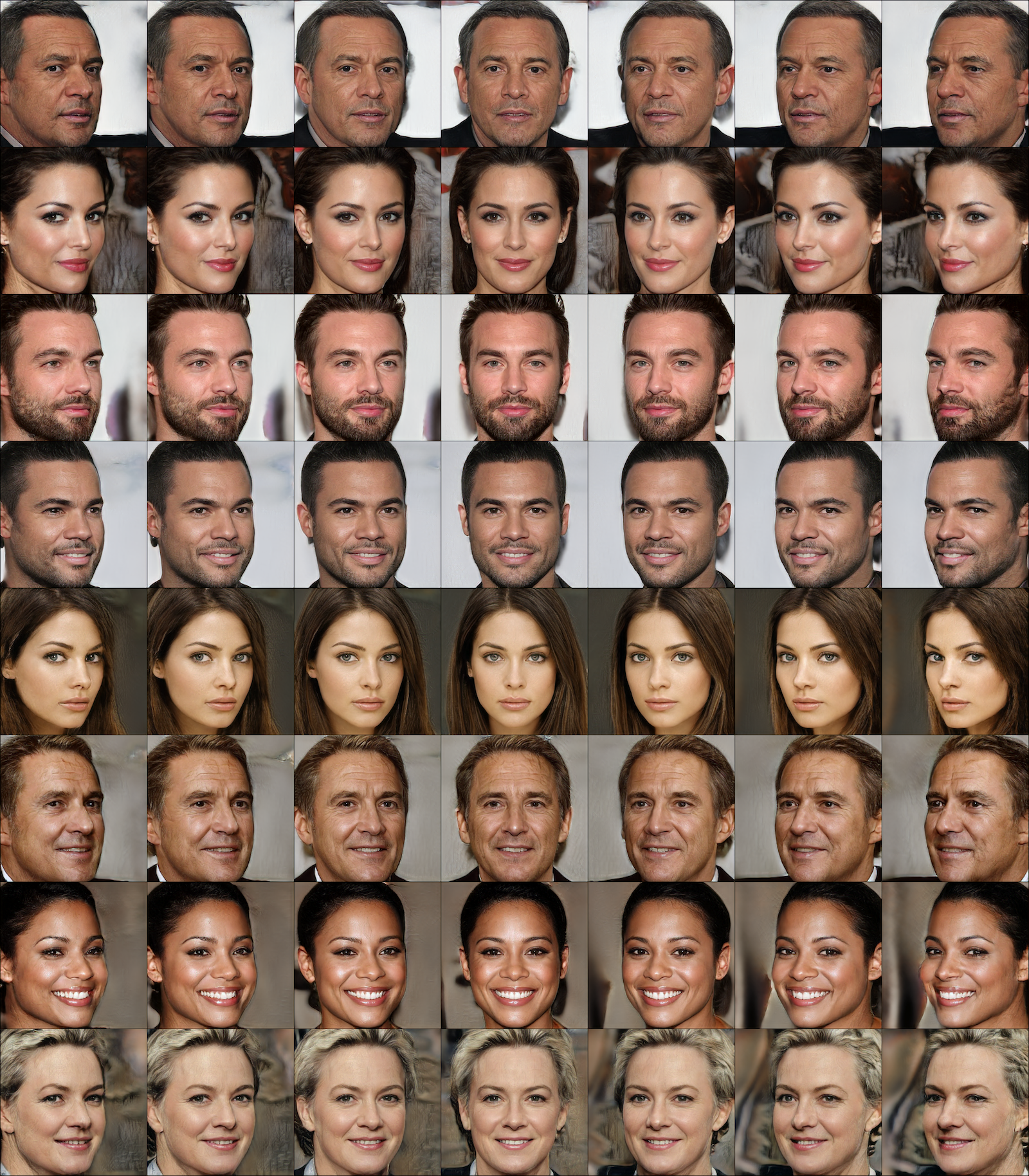}
  \caption{Images synthesized by MVCGAN on CELEBA-HQ~\cite{karras2018progressive} at resolution $512^2$.
}
\label{fig:cehq_sup}
\end{figure*}
\begin{figure*}[]
\centering
  \includegraphics[width=1 \linewidth]{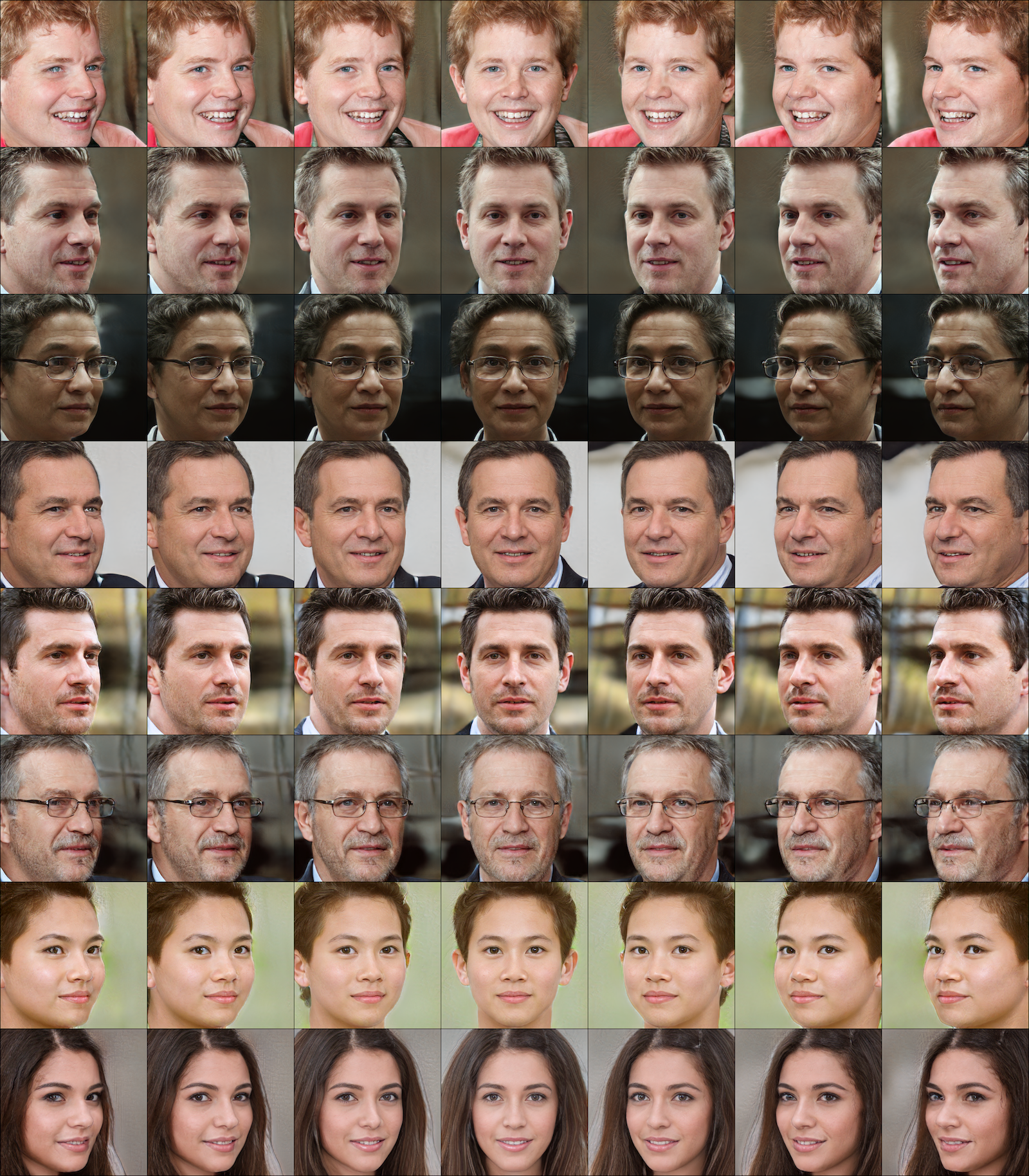}
  \caption{Images synthesized by MVCGAN on FFHQ~\cite{karras2019style} at resolution $512^2$.
}
\label{fig:ffhq_sup}
\end{figure*}

\subsection{Datasets}
\label{sec:data}
We conduct experiments on three widely-used high-resolution image datasets: CELEBA-HQ~\cite{karras2018progressive}, FFHQ~\cite{karras2019style}, and AFHQv2~\cite{choi2020stargan}.

\noindent \textbf{CELEBA-HQ.}
CELEBA-HQ\footnote{\url{https://github.com/tkarras/progressive_growing_of_gans}}~\cite{karras2018progressive} consists of 30,000 high-quality images of human face at $1024^2$ resolution.
During training, we sample the pitch and yaw of the camera pose from a Gaussian distribution with the horizontal standard deviation of 0.3 radians and the vertical standard deviation of 0.155 radians.

\noindent \textbf{FFHQ.}
Flickr-Faces-HQ~(FFHQ)\footnote{\url{https://github.com/NVlabs/ffhq-dataset}}~\cite{karras2019style} is a large scale human face dataset which contains 70,000 high-quality images at $1024^2$ resolution.
The images contain various styles with different ages, ethnicity, and background.
Besides, the humans in the images wear different accessories such as earrings, sunglasses, hats, and eyeglasses.
In the training stage, we sample the pitch and yaw of the camera pose from a Gaussian distribution with the horizontal standard deviation of 0.3 radians and the vertical standard deviation of 0.155 radians.

\noindent \textbf{AFHQv2.}
Animal Faces-HQ~(AFHQv2)\footnote{\url{https://github.com/clovaai/stargan-v2}}~\cite{choi2020stargan} contains 15,000 high-quality animal face images at $512^2$ resolution.
The dataset has three categories: cat, dog, and wildlife, with each category providing 5,000 images.
Following previous works~\cite{schwarz2020graf,niemeyer2021giraffe,chan2021pi},
we conduct experiments on the cat face images to make a fair comparison. 
During training, we sample the pitch and yaw of the camera pose from a uniform distribution with the horizontal standard deviation of 0.4 radians and the vertical standard deviation of 0.2 radians.

\subsection{Competitive Methods}
\label{sec:base}
We compare our approach against three state-of-the-art 3D-aware image synthesis methods: GRAF~\cite{schwarz2020graf}, pi-GAN~\cite{chan2021pi}, and GIRAFFE~\cite{niemeyer2021giraffe}.

\noindent \textbf{GRAF.}
We use the official implementation\footnote{\url{https://github.com/autonomousvision/graf}}
to train the model on CELEBA-HQ~\cite{karras2018progressive}, FFHQ~\cite{karras2019style} and AFHQv2~\cite{choi2020stargan} datasets.

\noindent \textbf{pi-GAN.}
We adopt the author's implementation\footnote{\url{https://github.com/marcoamonteiro/pi-GAN}} of pi-GAN~\cite{chan2021pi}.
Following the practice in pi-GAN~\cite{chan2021pi}, we begin training at $32^2$ and gradually increase to $128^2$ during training.
The high-resolution images are rendered by sampling rays more densely~($256^2$, $512^2$).

\noindent \textbf{GIRAFFE.}
We train GIRAFFE~\cite{karras2019style} on all datasets~\cite{karras2018progressive,karras2019style,choi2020stargan} with the official implementation\footnote{\url{https://github.com/autonomousvision/giraffe}}.

\section{Additional Results} \label{sec:add}
We provide additional results to show the multi-view consistency and the quality of the generated images.

\noindent \textbf{3D Reconstruction.}
To further demonstrate the multi-view consistency of our method, we recover the 3D shape from  generated images  with the 3D reconstruction method~\cite{schonberger2016structure}.
As shown in Fig.~\ref{fig:full_view}, we render images of a single instance from 35 views, and then perform dense 3D reconstruction by running COLMAP~\cite{schonberger2016structure} with default parameters and no known camera poses.
The results in Fig.~\ref{fig:colmap} validate the correctness of the 3D shape learned by our model.

\noindent \textbf{More Visualization Results.}
We provide more generated images in Fig.~\ref{fig:cehq_sup} and Fig.~\ref{fig:ffhq_sup}.
Please also refer to the supplementary video for more results.


\end{document}